\documentclass[conference]{IEEEtran}

\usepackage{cite}

\usepackage[pdftex]{graphicx}
\graphicspath{{_img/}}
\DeclareGraphicsExtensions{.pdf,.jpeg,.png}

\usepackage{algpseudocode}

\begin{document}
%
\title{Cortical Processing with Thermodynamic-RAM}

\author{
  \IEEEauthorblockN{M. Alexander Nugent}
  \IEEEauthorblockA{M. Alexander Nugent Consulting, Santa Fe, NM, USA\\
  KnowmTech LLC, Albuquerque, NM, USA\\
  Xeiam LLC, Santa Fe, NM, USA\\
  Email: i@alexnugent.name}
  \and
  \IEEEauthorblockN{Timothy W. Molter}
  \IEEEauthorblockA{M. Alexander Nugent Consulting, Santa Fe, NM, USA\\
  KnowmTech LLC, Albuquerque, NM, USA\\
  Xeiam LLC, Santa Fe, NM, USA\\
  Email: tim.molter@gmail.com}
}

\maketitle

\begin{abstract}

AHaH computing forms a theoretical framework from which a biologically-inspired type of computing architecture can be built where, unlike von Neumann systems, memory and processor are physically combined. In this paper we report on an incremental step beyond the theoretical framework of AHaH computing toward the development of a memristor-based physical neural processing unit (NPU), which we call Thermodynamic-RAM (kT-RAM). While the power consumption and speed dominance of such an NPU over von Neumann architectures for machine learning applications is well appreciated, Thermodynamic-RAM offers several advantages over other hardware approaches to adaptation and learning. Benefits include general-purpose use, a simple yet flexible instruction set and easy integration into existing digital platforms. We present a high level design of kT-RAM and a formal definition of its instruction set. We report the completion of a kT-RAM emulator and the successful port of all previous machine learning benchmark applications including unsupervised clustering, supervised and unsupervised classification, complex signal prediction, unsupervised robotic actuation and combinatorial optimization. Lastly, we extend a previous MNIST hand written digits benchmark application, to show that an extra step of reading the synaptic states of AHaH nodes during the train phase (healing) alone results in plasticity that improves the classifier's performance, bumping our best F1 score up to 99.5\%.

\end{abstract}

\IEEEpeerreviewmaketitle

\section{Introduction}

Because of its superior power, size and speed advantages over traditional von Neumann computing architectures, the biological cortex is one of Nature's structures that humans are trying to understand and recreate in silicon form \cite{mccormick2014applying,navaridas2013spinnaker,benjamin2014neurogrid,esser2013cognitive}. The cognitive ability of causal understanding that animals possess along with other higher functions such as motor control, reasoning, perception and planning are credited to different anatomical structures in the brain and cortex across animal groups \cite{dugas2012cell}. While the overall architecture and neural topology of the cortex across the animal kingdom may be different, the common denominator is the neuron and its connections via synapses. 

In this paper, we present a general-purpose neural processing unit (NPU) hardware circuit we call Thermodynamic-RAM that provides a physical adaptive computing resource to enable efficient emulation of functions associated with brains and cortex. The motivation for this work is to satisfy the need for a chip where processor and memory are the same physical substrate so that computations and memory adaptations occur together in the same location. For machine learning applications that intend to achieve the ultra-low power dissipation of biological nervous systems, the shuttling of bits back and forth between processor and memory grows with the size and complexity of the model. The low-power solution to machine learning occurs when the memory-processor distance goes to zero and the voltage necessary for adaption is reduced. This can only be achieved through intrinsically adaptive hardware.

Other experts in the field have also expressed the need for this new type of computing. Traversa and Ventra recently introduced the idea of `universal memcomputing machines', a general-purpose computing machine that has the same computational power as a non-deterministic Universal Turing Machine but also exhibiting intrinsic parallelization and information overhead \cite{traversa2014universal}. Their proposed DCRAM and other similar solutions employ memristors, elements that exhibit the capacity for both memory and processing. They show that a memristor or memcapacitor can be used as a subcomponent for computation while at the same time storing a unit of data. A previous study by Thomas et al. also argued that the memristor can better be used to implement neuromorphic hardware than traditional CMOS electronics \cite{thomas2013memristor}. In recent years there has been an explosion of worldwide interest in memristive devices \cite{prodromakis2010review}, their connection to biological synapses \cite{chang2011short,li2013ultrafast}, and use in alternative computing architectures \cite{morabito2013neuromorphic}.

Thermodynamic-RAM is adaptive hardware operating on the principles of `AHaH computing' \cite{nugent2014ahah}, a new computing paradigm where processor and memory are united. The fundamental building blocks it provides are `AHaH nodes' and `AHaH synapses' analogous to biological neurons and synapses respectively. An AHaH node is built up from one or more synapses, which are are implemented as differential memristor pairs. Spike streams asynchronously drive co-activation of synapses, and kT-RAM's instruction set allows for specification of supervised or unsupervised adaptive feedback. The co-active synaptic weights are summed on the AHaH node's output electrode as an analog sum of currents providing both a state and a magnitude. While reverse engineering the biological brain was not the starting point for developing AHaH computing, it is encouraging that the present design shares similarities with the cortex such as sparse spike data encoding and the capacity for continuous on-line learning.

Much like a graphical processing unit (GPU) accelerates graphics, kT-RAM plugs into existing computer architectures to accelerate machine learning operations. This opens up the possibility to give computer hardware the ability to perceive and act on information flows without being explicitly programmed. In neuromorphic systems, there are three main specifications: the network topology, the synaptic values and the plasticity of the interconnections or synapses. Any general-purpose neural processor must contend with the problem that hard-wired neural topology will restrict the available neural algorithms that can be run on the processor. In kT-RAM, the topology is defined in software, and this flexibility allows for it to be configured for different machine learning applications requiring different network topologies such as trees, forests, meshes, and hierarchies. Much like a central processing unit (CPU) carries out instructions of a computer program to implement any arbitrary algorithm, kT-RAM is also general-purpose, in that it does not enforce a specific network topology. Additionally, a simple instruction set allows for various forms of synaptic adaptation, each useful within specific contexts. So far as we currently understand, the kT-RAM instruction set is capable of supporting universal machine learning functions.

For an exhaustive introduction to AHaH computing, AHaH nodes, the AHaH rule, and the AHaH node as well as circuit simulation results, machine learning benchmark applications, and accompanying open source code, please refer to a previous paper entitled ``AHaH computing–From Metastable Switches to Attractors to Machine Learning'' \cite{nugent2014ahah}.

\section{Methods}

\subsection{Architecture}

The architecture of Thermodynamic-RAM presented in this paper is a particular design that prioritizes flexibility and general utility above anything else, much in the same way that a CPU is designed for general purpose use. Different machine learning applications require different network topologies (the way in which neurons are connected to each other), and having a chip that can be configured for any desired network topology has the broadest general appeal across the field. Our kT-RAM design borrows heavily from commodity RAM using its form factor and the row and column address space mapping to specific bit cells as a basis to build upon. Converting RAM to kT-RAM requires the following steps: (1) the removal of the RAM reading circuitry, (2) minor design modifications of the RAM cells, (3) the addition of memristive synapses to the RAM cells, (4) the addition of H-Tree circuitry connecting the synapses (5) and the addition of driving and output sensing circuitry.

\begin{figure}[!t]
\centering
\includegraphics[width=1.0\linewidth]{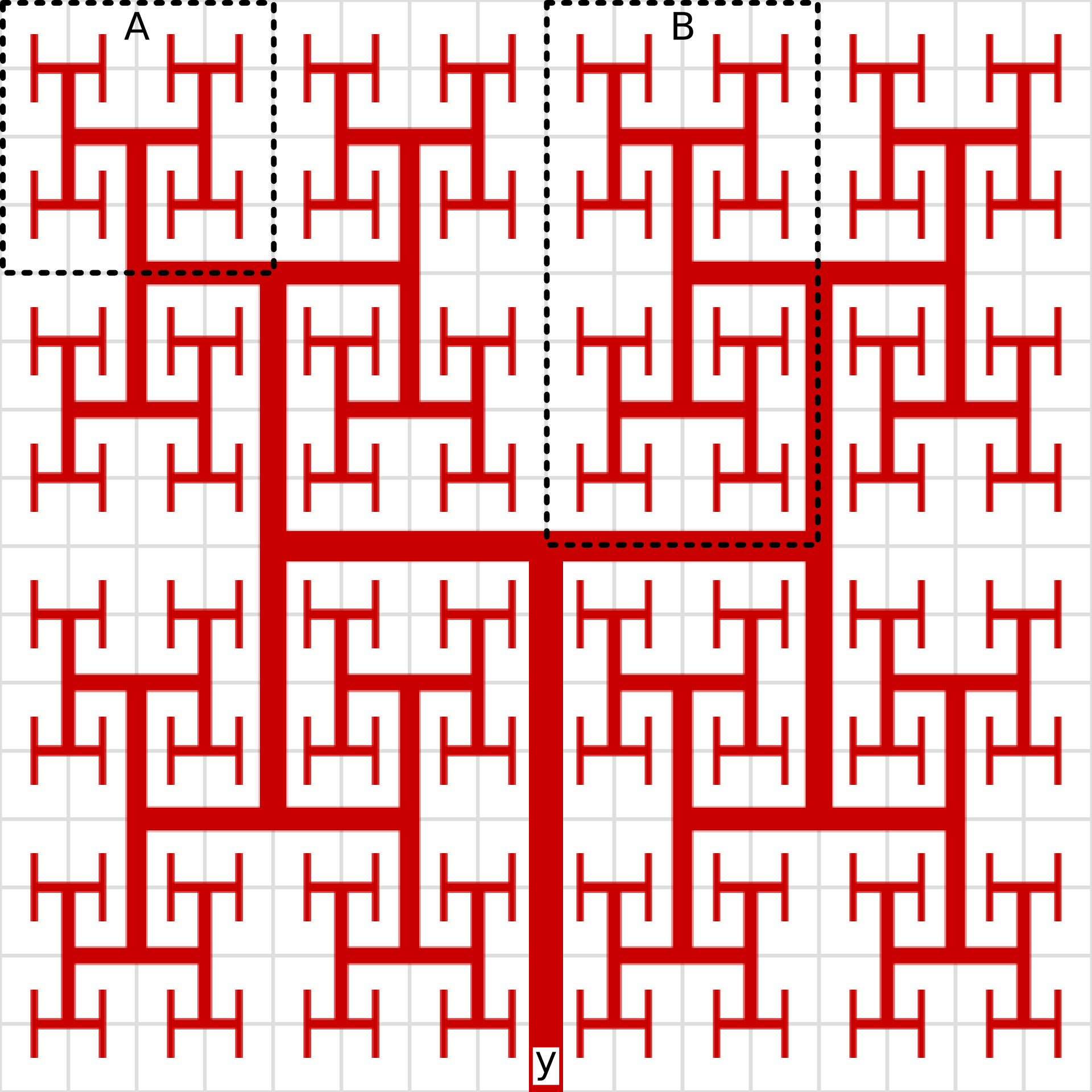}
\caption{Thermodynamic-RAM utilizes standard RAM technology for synaptic activation over a two-dimensional address space (light gray-bordered cells). The fractal H-Tree wire (red) forms a common electrical node (y) for summing the synaptic weights of an AHaH node (neuron) and also for providing a learning/feedback signal. Here an 16 x 16 cell array is shown, but in practice a much larger array containing many thousands of synapses can be fabricated. AHaH node temporal partitioning is achieved by addressing different spaces at different points in time (AHaH nodes A and B).}
\label{fig_htree}
\end{figure}

Figure \ref{fig_htree} shows an example of what kT-RAM would look like with its H-Tree sensing node on the top metal layers of the chip connecting all the underlying synapses located at each cell in the RAM array. While at first glance it appears like this architecture leads to one giant AHaH node per chip or core, the core can be partitioned into smaller AHaH nodes of arbitrary size by temporally partitioning sub portions of the tree (AHaH nodes A and B for example). All other synapses that are not included in the currently-focused AHaH node space are unaffected by any read or write operations as they are floating. In other words, so long as it is guaranteed that synapses assigned to a particular AHaH node are never co-activated with other AHaH node partitions, these `virtual' AHaH nodes can co-exist on the same physical core. Any desired network topology linking AHaH nodes together can be achieved by this temporal partitioning concept by utilizing standard RAM to store network topologies and core partitions. Software enforces the constraints, while the hardware remains flexible. It is worth mentioning that many real-world applications of machine learning do not cleanly fit within the idealized structure of a pure network. Rather, mixed computation and algorithms are often used. By pairing kT-RAM with a standard CPU and RAM, enormous flexibility in use is possible. 

\begin{figure}[!t]
\centering
\includegraphics[width=1.0\linewidth]{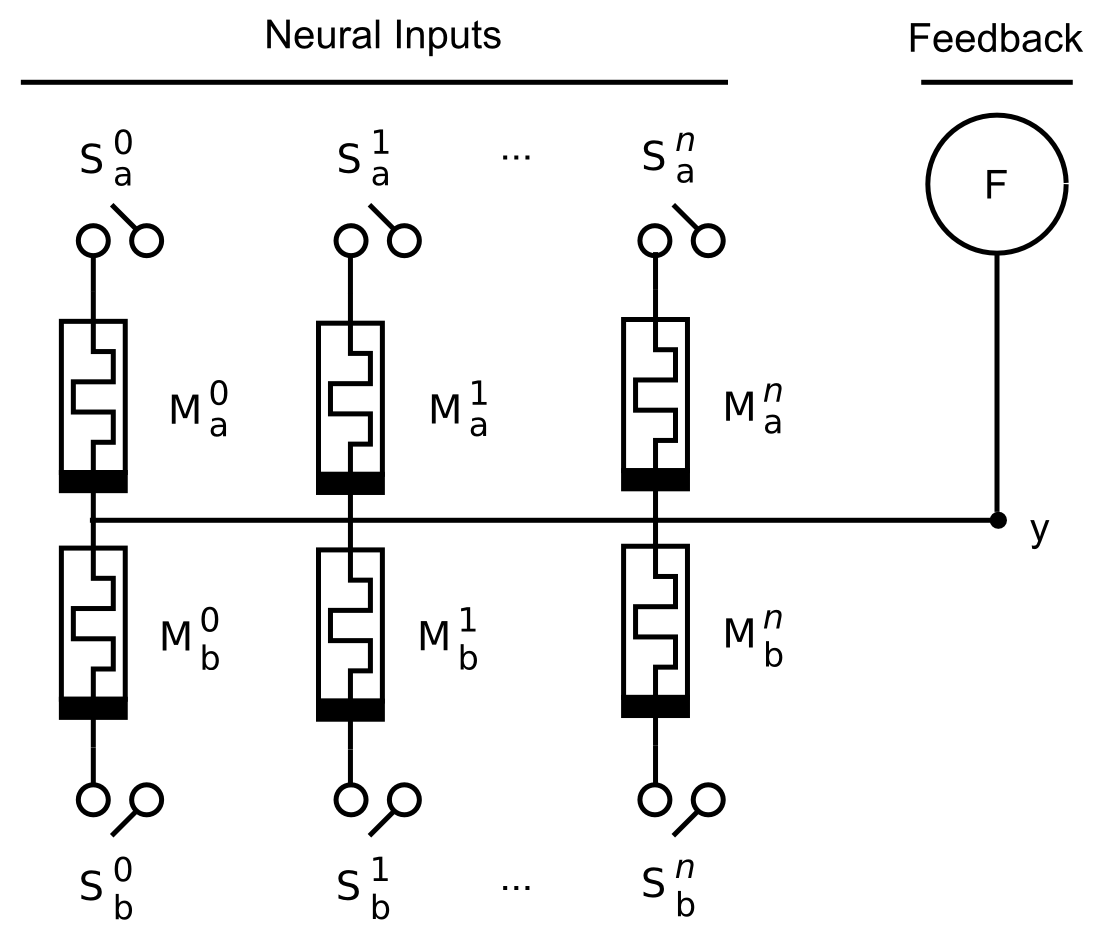}
\caption{An AHaH node is the basic building block of AHaH computing. Thermodynamic-RAM provides a physical substrate for forming AHaH nodes (neurons) made up of serially-connected memristor pairs (synapses). The individual neural inputs corresponding to an incoming sparse-encoded spike stream are activated using standard RAM address mapping. The common electrode labeled `y' serves as both a node for summing the weights of the activated synapses and also for delivering learning feedback to the synapses.}
\label{fig_ahah_node}
\end{figure}

Figure \ref{fig_ahah_node} shows a model of an AHaH node as implemented with switches and memristors. The AHaH node has one common node linking $n$ synapses, with $k$ active inputs and $n-k$ inactive (floating) inputs at any given time. The individual gated switches for activating spike inputs are labeled ${\rm S}^0$, ${\rm S}^1$, $\cdots$ ${\rm S}^n$. Each synapse is made up of two serially-connected memristors labeled ${\rm M}^0$, ${\rm M}^1$, $\cdots$ ${\rm M}^n$. The driving voltage source for supervised and unsupervised learning is labeled F. The subscript values a and b indicate the positive and negative dissipative pathways, respectively.

\subsection{Thermodynamic-RAM Emulator}

The substantial cost of hardware, let alone the cost of new hybrid memristor-CMOS hardware, provides tremendous inhibitory pressure against the realization of an NPU. On the one hand an NPU design needs to justify its existence by demonstrating utility across one or more application spaces. On the other hand, one cannot develop applications without the hardware. We believe we have found a solution to this chicken-and-egg problem in the form of a digital kT-RAM emulator. The AHaH node is very simple, and we have exploited this simplicity to create an efficient digital emulator. 

The digital emulator emulates simplified AHaH nodes where the memristors are represented as having a set of incremental conductance states. The `nibble core', for example, treats each memristor as having 16 discrete states and can store a synapse with one byte. The `byte core' models memristors as having 256 discrete states and can store a synapse with two bytes. The `float core' uses floating-point precision and provides a point of correspondence to our analog emulator. The analog emulator is more computationally expensive to run as it emulates a circuit containing memristors which behave according to our `generalized metastable switch memristor model' \cite{nugent2014ahah}. The memristors which we have previously applied our model to include the Ag-chalcogenide \cite{oblea2010silver}, AIST \cite{zhang2013aginsbte}, GST \cite{li2013ultrafast}, and WO$_x$ \cite{chang2011synaptic} devices. We have not changed the model's parameters during the integration of the memristors into the analog emulator since our previous study \cite{nugent2014ahah}. Node voltages are solved using Kirchoff's Current Law (KCL). RC delays and capacitive effects of switching have been ignored for now. It is worth noting that all above memristor devices have proven to produce similar performance scores on the benchmarks amenable to using the analog emulator. In general, as long as the memristor is incremental and its state change is voltage dependent, it should be a viable candidate for kT-RAM. 

Application developers can commence with building a market for kT-RAM on existing hardware platforms (smart phones, servers, etc.) while hardware developers can work to build more accurate and detailed emulators. Importantly, hardware developers can look to the application developers to see where the application spaces actually are and what is, and is not, useful in the real world. Although memristive kT-RAM is the long term goal, multiple generations of pure digital kT-RAM emulators can be built, each offering various trade-offs in terms of bandwidth, memory and power. Through adherence to the kT-RAM instruction set, programs can be ported from one technology generation to another and multiple industry participants can coordinate their actions across the whole technology stack. Developments at the hardware level can be informed by the application level, where utility is demonstrated, and innovations at the hardware level can propagate back to the application level.

\subsection{Thermodynamic-RAM Instruction Set and Operation}

As stated previously, Thermodynamic-RAM can be plugged into existing digital computing architectures. The envisioned hardware format is congruent with standard RAM chips and RAM modules and would plug into a motherboard in a variety of different ways. In general there are two main categories of integration. First, kT-RAM is tightly coupled with the CPU, on the CPU die itself or connected via the north bridge. In this case, the instruction set of the CPU would have to be modified to accommodate the new capabilities of kT-RAM. The CPU would communicate with the kT-RAM module in a similar manner as with normal RAM, over an address and data bus. Whereas normal RAM returns a block of bits, kT-RAM would return a digitized value of the AHaH node's analog voltage on electrode y, shown in Figure \ref{fig_ahah_node}. Alternatively, kT-RAM is loosely coupled as a peripheral device either connected via the PCI bus, the LPC bus, or via cables or ports to the south bridge. In these cases, no modification to the CPU's instruction set would be necessary, as the interfacing would be implemented over the generic plug-in points of the south bus. As in the case with other peripheral devices, a device driver would need to be developed. Additional integration configurations are also possible.

Given the above hardware integration, kT-RAM simply becomes an additional resource that software developers have access to via an API. Currently kT-RAM is implemented as an emulator running on von Neumann architecture. Later, when a chip is available, it will replace the emulator, and existing programs written against the API will not need to be rewritten to benefit from the accelerated capabilities offered by the new hardware. In any case, kT-RAM or emulated kT-RAM operates asynchronously. As new spikes arrive, the driver in control of kT-RAM is responsible for activating the correct synapses and providing an instruction pair for each AHaH node. The returned activation value can then be passed back to the program and used as needed. 

The following steps are carried out by the system to process spike streams: (1) sequentially load all active synapses corresponding to the spike pattern, (2) provide an instruction or instruction pair (read-write), (3) kT-RAM executes the supplied read instruction, (4) receive the AHaH node's activation value (confidence) from kT-RAM and (5) kT-RAM executes the supplied write instruction (if provided).

One can theoretically exploit the kT-RAM instruction set (Table \ref{instruction_set}) however they wish. However, to prevent saturation of the memristors in a maximally- or minimally-conductive state, one must pair `forward' instructions with `reverse' instructions. For example, a forward-read operation $FF$ should be followed by a reverse operation ($RF$, $RH$, $RL$, $RZ$, $RA$ or $RU$) and vise versa. The only way to extract state information is to leave the feedback voltage floating, and thus there are two possible read instructions: $FF$ and $RF$. There is no such thing as a `non-destructive read' operation in kT-RAM. Every memory access results in weight adaptation according to AHaH plasticity. By understanding how the AHaH rule works (AHaH computing), we can exploit the weight adaptations to create, among other things, `self-healing hardware'. 

\begin{table}[!t]
\renewcommand{\arraystretch}{1.3}
\caption{kT-RAM Instruction Set}
\label{instruction_set}
\centering
\begin{tabular}{|c|c|c|c|c|c|c|}
\hline
Instruction & Synapse Driving Voltage & Feedback Voltage (F)\\
\hline
FF & Forward-Float & None/Floating \\
\hline
FH & Forward-High & $-V$ \\
\hline
FL & Forward-Low & $+V$ \\
\hline 
FU & Forward-Unsupervised & $-V$ if $y\geq0$ else $+V$ \\
\hline
FA & Forward-Anti-Unsupervised & $+V$ if $y\geq0$ else $-V$ \\
\hline
FZ & Forward-Zero & $0$ \\
\hline
RF & Reverse-Float & None/Floating \\
\hline
RH & Reverse-High & $-V$ \\
\hline
RL & Reverse-Low & $+V$ \\
\hline
RU & Reverse-Unsupervised & $-V$ if $y\geq0$ else $+V$ \\
\hline
RA & Reverse-Anti-Unsupervised & $+V$ if $y\geq0$ else $-V$ \\
\hline
RZ & Reverse-Zero & $0$ \\
\hline
\end{tabular}
\end{table}

\subsection{Thermodynamic-RAM Classifier}

Figure \ref{fig_classifier} contains pseudo code demonstrating how to construct a multi-label online classifier in software by loading spikes and executing instructions in the kT-RAM instruction set. The network topology of the classifier is simply $N$ AHaH nodes with $M$ synapses, where $N$ is the number of labels being classified and $M$ is the number of unique spikes in the entire spike stream space. The active spikes $S$, a subset of $M$, is loaded onto each AHaH, and the execute method returns the voltage on the AHaH node's output electrode, y. Although all the AHaH nodes may exist on the same physical chip and share the same output electrode, temporal partitioning, as described above, allows for a virtual separation of AHaH nodes.

\begin{figure}
\begin{algorithmic}[1]
\Procedure{Classify}{active spikes set $S$, truth labels set $L$}
\For{Each AHaH Node $N$}
\State KTRAM.loadSpikes($N$, $S$)
\State $y\gets$ KTRAM.execute($N$, $FF$) \Comment{forward read}
\If{supervised}

\If{$N \in L$}
\State KTRAM.execute($N$, $RH$)
\ElsIf{$y \geq 0$}\Comment{false-positive}
\State KTRAM.execute($N$, $RL$)
\Else \Comment{true-negative}
\State KTRAM.execute($N$, $RF$) 
\EndIf

\Else \Comment{unsupervised}
\State KTRAM.execute($N$, $RF$) 
\EndIf
\EndFor
\EndProcedure
\end{algorithmic}
\caption{A multi-label online linear classifier with confidence estimation can be easily constructed via calls to the kT-RAM instruction set.}
\label{fig_classifier}
\end{figure}

\subsection{MNIST Benchmark}

The Mixed National Institute of Standards and Technology (MNIST) database \cite{lecun1998mnist} is a classic dataset in the machine learning community. It consists of 60,000 train and 10,000 test samples of handwritten digits, each containing a digit 0 to 9 (10 classes). The 28 x 28 pixel grayscale images have been preprocessed to size-normalize and center the digits.

Our approach to the MNIST benchmark is to couple decision trees acting as spike encoders to a back-end classifier. We have ported multiple variants of adaptive and non-adaptive decision trees, as well as the classifier, to the kT-RAM emulator. For the work presented here we use both a simple non-adaptive and efficient decision tree to reduce the computational load as well as the same adaptive decision tree used in \cite{nugent2014ahah}. Our choice of the non-adaptive random decision tree is also motivated by the desire to isolate adaptation only to the classifier while studying the effects of healing. Previously, AHaH attractor states have been shown to be computationally complete \cite{nugent2014ahah}, which means that a decision tree for spike encoding formed of AHaH nodes in various configurations is capable of performing a very large number of possible feature encodings. We have only just begun exploring the space.

The classification procedure is to (1) spike-encode the data and (2) perform a multi-label classification of the resulting spike stream. The mechanisms of the spike encoding strongly effect the resulting classification performance, both in terms of error rate but also computational efficiency. The optimal spike encoding method is determined by the problem constraints. The encoding can be done automatically or manually. The former entails feature learning or selection, and the latter entails feature engineering. 

We first create binary representations of each image by thresholding pixels exceeding a value of ten. The output of this thresholding over an image patch of 8 x 8 pixels produces a spike stream with a space of 64 channels. This spike stream is in turn fed into a simple random spike decision tree, where each node in the tree is looking for the presence of a spike on one spike channel. The leaves of the decision tree encode features, and these `feature spike streams' were joined with pooling coordinates to produce a final output spike stream that was fed to the classifier. Each spike in the final spike stream represents the presence of a feature in a non-overlapping pooling region (8 x 8 pixels). This architecture could be compared roughly to a simple convolutional neural network. 

The output of the classifier produces a list of AHaH node `activations' in the form of voltages. The more positive the voltage, the more likely that it is correct. By increasing a confidence threshold, one can increase the classification precision at the expense of reducing recall. Since the choice of confidence threshold is arbitrary, we report here the peak F1 score.

In the absence of any training labels, each AHaH node in the classifier receives an $FF-RF$ instruction sequence, which amounts to a forward voltage to read the node state followed by a reverse voltage without feedback. This instruction sequence is carried out during the train phase of the classification and the $RF$ operation is necessary to prevent the memristors from saturating. Depending on the initial state of the synapses and the evaluation voltage, the synapses will slightly adapt, meaning there is no possible way to perform a non-destructive read operation. While one might assume that a read operation which adapts the memristors' state is detrimental, this adaptation actually improves the classifier's performance indicating that a slight dose of online unsupervised learning occurs by just reading. 

This mechanism can be taken advantage of to further improve the results of the MNIST classification benchmark by performing a series of unsupervised classifications on subsets of the original spike stream during the training phase (healing). We extended the baseline classifier procedure as shown in Figure \ref{fig_classifier} so that during the train phase an extra classification was performed. For each training exemplar, the base supervised classification was performed followed by another classification with a randomly-chosen subset of the original spike stream's activated spikes. We varied this amount from 0 to 100\%. We also tested two variations of this, where the spike pattern subset was reclassified either via `unsupervised' ($FF-RF$) or `supervised' (see Figure \ref{fig_classifier}). Each configuration was repeated 10 times, and each experiment included 3 training epochs over a reduced set of 10,000 training and 1,000 test images. The reduced train and test set was motivated by the need to speed up total experiment time. As stated above, a non-adaptive spike encoder was used for this experiment in order to isolate the adaptive healing effects to only the classifier. 

As a final experiment, we repeated our original MNIST classification benchmark with the full train and test image set, an adaptive spike encoder, and the new healing method.

\section{Results and Discussion}

Here, we report that all the benchmarks and challenges from the previous AHaH computing publication \cite{nugent2014ahah} were successfully re-evaluated using the kT-RAM emulator. The complete list includes visual clustering challenges, unsupervised robotic arm challenge, 64-city traveling-salesman problem, complex signal prediction challenge, Breast Cancer Wisconsin, Reuters-21578 Distribution 1.0, and Census Income classification benchmarks \cite{Bache+Lichman:2013}. All benchmarks and challenges performed roughly the same as before using the kT-RAM emulator with the various core types. As expected, the low-resolution nibble and byte cores, while extremely fast, cause slight degradation of performance. This trade off was observed with the higher resolution core types as well, where their performance was better at the cost of longer processing times. The analog core, which models a particular Ag-chalcogenide memristor from Boise State University showed excellent congruence with previous benchmarks, and also exhibited the expected speed degradation given the extra complexity of the model.

\subsection{MNIST Benchmark}

\begin{figure}[!t]
\centering
\includegraphics[width=1.0\linewidth]{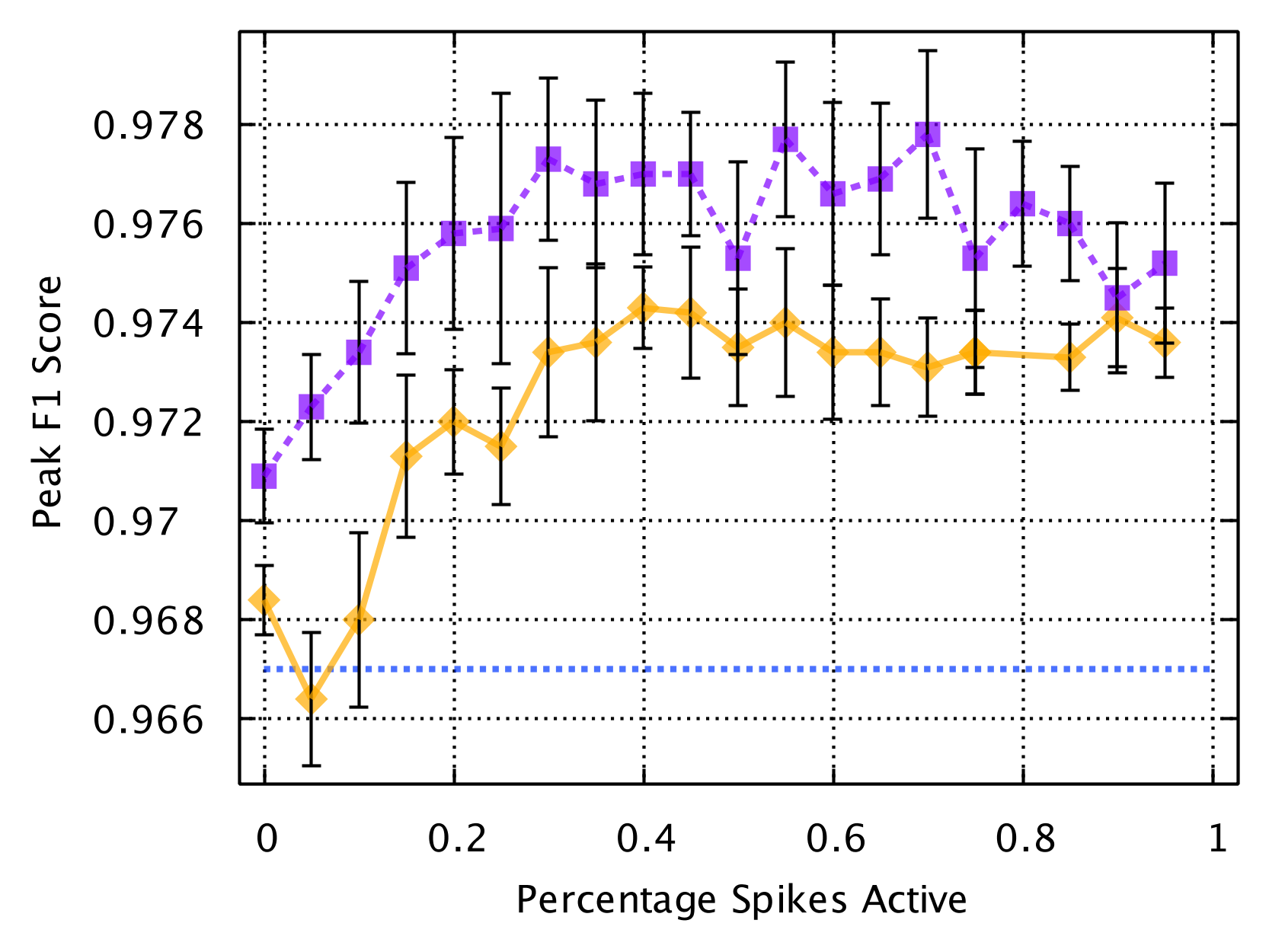}
\caption{For each training exemplar in the MNIST classification benchmark, a non-adaptive spike encoder was used to generate a spike encoded input to the kT-RAM classifier. Random subsets of the original spike-encoded input (from 0\% to 95\%) were generated from the original spike pattern and reclassified (healing), either unsupervised (purple, dashed) or supervised (orange). Average baseline F1 score for 3 epochs was 0.967 (blue, dotted). The results demonstrate that the act of reading the synaptic states of AHaH nodes results in plasticity that improves the classifier's performance on the test set.}
\label{fig_sweep}
\end{figure}

As shown in Figure \ref{fig_sweep}, the baseline classification F1 score for the non-adaptive spike encoder, without healing, and with 10,000 training images and 1,000 test images was 0.967. The degradation of F1 score is due to the reduced train-test set. It can been seen that reclassifying a subset of each exemplar's spike set during the train phase (healing) improves the results beyond the baseline. Furthermore, the unsupervised version outperformed the supervised version. In general a percentage of active spikes used for the reclassification of 30\% to 70\% gave the best results. The kT-RAM classifier is therefore capable of self-optimizing its performance in an unsupervised way. Furthermore, this demonstrates that read operations do not adversely affect the classification performance but rather improve them. In other words, there is no need for a non-destructive read operation, which is commonly assumed by circuit designers. This opens up the possibility of ultra-low power adaptive circuitry, since adaptation can theoretically occur at low voltages and the circuit can heal itself from the damage that will occur due to thermal fluctuations.

Repeating the MNIST classification benchmark on the full train and test dataset (60,000 and 10,000 images respectively) using our new kT-RAM emulator, the original adaptive spike encoder and healing yielded better performance than our previous results of 0.98 - 0.99 F1 score \cite{nugent2014ahah}. Using the adaptive encoder of \cite{nugent2014ahah} over 20 training epochs yielded an F1 score of 0.995, taking approximately 2.5 hours to train running on a 2012 MacBook Pro. These results show that our port of AHaH computing functional code over to our new kT-RAM emulator code did not degrade the outcome of the classifier. Just adding the healing mechanism significantly improved results. In addition, the experiment time actually decreased, indicating that our emulator is very efficient.

\section{Conclusion}

In this paper, we have proposed one possible hardware implementation of AHaH computing, which we call Thermodynamic-RAM or kT-RAM for short. While a detailed description of the chip design is beyond the scope of this paper, an overview of how standard RAM can be modified to create kT-RAM was given. Additionally, we described how kT-RAM can be plugged into existing digital computing platforms in a similar way as RAM and RAM modules. Whether kT-RAM is integrated directly into a CPU's die, or it is off-board as a peripheral device or anything in between, the kT-RAM instruction set provides a minimal and complete interface for integrating the chip into existing systems for accelerating machine learning tasks where large scale adaptive networks become a bottleneck in typical von Neumann architecture. Writing software to utilize kT-RAM will only require the addition of the kT-RAM API. Our kT-RAM emulator allows us to develop applications, demonstrate utility, and justify a large investment into chip development. When chips are available, existing applications using the emulator API will not have to be rewritten in order to take advantage of new hardware acceleration capabilities. 

In addition, we report that we have successfully ported all of our previous examples of AHaH machine learning capabilities to use our new kT-RAM emulator including classification, prediction, clustering, robotic control, and combinatorial optimization. All of these capabilities are are associated with biological cortex. The computing architecture in both cortex and kT-RAM is one in which the memory and processor are the same physical substrate. The temporal partitioning feature of the kT-RAM design presented allows for a general purpose chip where the network topology is configured in software rather than being `hard coded' in the chip's circuitry. The software procedure for implementing a classifier with kT-RAM and the instruction set was given as pseudo code, and it shows how the spike streams and truth labels (for supervised learning) are processed. Extending our previous MNIST hand written digit classification benchmark, it was shown that the act of reading the synaptic states of AHaH nodes (healing) alone results in plasticity that improves the classifier's performance. Our previous best F1 score on the MNIST classification benchmark has been improved to 0.995. Note that it is not our intention to produce the absolute best MNIST benchmark performance and compete with established machine learning techniques. We are simply choosing a wide range of machine learning benchmarks and challenges to show utility of our proposed NPU. The advantage will be low-power adaptive hardware with a consistent API that can be used to solve numerous problems across the wide spectrum of the machine learning application space. What we are attempting to present is therefore not just a specific solution to a benchmark problem but rather that kT-RAM offers a path toward a new type of adaptive generic computing substrate. 

\section{Future Work}

The classifier demonstration in this paper represents just one network topology and one basic capability of the cortex, but as mentioned, all the machine learning capabilities reported in a previous publication \cite{nugent2014ahah} have been successfully ported over to the kT-RAM emulator. Different topologies implemented by different procedures and via the described AHaH node temporal partitioning can be utilized for a wide range of cortical capabilities. Because the network topology is defined in software and not `hard-coded' in kT-RAM circuitry, any topology can be created, explored and tested. This flexibility reminds one of the advantages and attractiveness of the CPU - it is a jack of all trades and master of none. This is not to say that kT-RAM could not be redesigned to an application-specific version, just as ASICs are optimized circuits that are designed to do one thing well. Substantial work remains in hardware design and fabrication and more generally in further developing AHaH computing. The simplicity of Thermodynamic-RAM, combined with the fact that we have demonstrated a number of machine learning capabilities leads us to conclude that this work will be well worth the effort.

\section*{Acknowledgment}

The authors would like to thank the Air Force Research Labs in Rome, NY for their support under the SBIR/STTR programs AF10-BT31, AF121-049. 


\bibliographystyle{IEEEtran}
%




\end{document}